\documentclass[conference,a4paper]{IEEEtran}
\usepackage[hyphens]{url}
\usepackage{graphicx}
\usepackage{booktabs}
\usepackage{amsmath}
\usepackage{cite}
\usepackage[hidelinks]{hyperref}
\hypersetup{
  pdftitle={Accuracy Hides How Language Models Fail: Measuring Failure States Under Matched Output Budgets},
  pdfauthor={Zongyou Yang; Yinghan Hou},
  pdfsubject={Two-layer measurement of language-model failure states},
  pdfkeywords={language model evaluation, failure states, output budgets, scorer provenance}
}

\title{Accuracy Hides How Language Models Fail: Measuring Failure States Under Matched Output Budgets}
\author{
\IEEEauthorblockN{Zongyou Yang}
\IEEEauthorblockA{\textit{Dyson School of Design Engineering}\\
\textit{Imperial College London}\\
London, United Kingdom\\
zy2926@ic.ac.uk}
\and
\IEEEauthorblockN{Yinghan Hou}
\IEEEauthorblockA{\textit{Department of Electrical and Electronic Engineering}\\
\textit{Imperial College London}\\
London, United Kingdom\\
yh24@ic.ac.uk}
}

\begin{document}
\maketitle

\begin{abstract}
Language-model benchmarks collapse two distinct measurement questions into a single accuracy score: whether a response reached an evaluable state, and whether its answer was judged correct. We introduce a two-layer evaluation framework that separates scorer-independent execution evidence---termination, answer exposure, parseability, and completion length---from scorer-dependent correctness. Across 2,550 outputs from five fixed Qwen and DeepSeek configurations on MATH and ARC-Challenge, matched 2,048-token limits produce sharply different execution mixtures: 49 of 450 Qwen MATH outputs terminate without a final answer, compared with 5 of 300 DeepSeek MATH outputs and none of the 750 ARC outputs. Among the same 300 DeepSeek MATH question--model pairs, no missing-final length termination is observed at 8,192 tokens. A coverage-audited targeted verification study further shows that candidate-selection and aggregation policies can substantially alter comparative accuracy estimates. These results demonstrate that accuracy conflates execution case mix with verification policy. Evaluations of test-time methods should therefore report pre-intervention execution states, verification coverage, and scorer provenance alongside accuracy.
\end{abstract}

\section{Introduction}

Accuracy compresses several observably different outcomes into one bit. A response can hit its output limit before presenting an answer, terminate normally without a parseable answer, or complete with a wrong answer. These cases expose different evidence to a test-time method: continuation is possible at a token boundary, whereas a completed wrong answer may require revision or verification. Aggregate recovery accuracy can therefore depend on the \emph{case mix} presented to a method, not only on the method itself.

Correctness is also a measurement outcome. Symbolic equivalence, option-label normalization, and model-based grading can move a parsed response between wrong and correct without changing its execution record. Prior work has documented both the usefulness and the biases of LLM evaluators \cite{liu2023geval,zheng2023judge,wang2024notfair,dubois2024lengthcontrolled,doddapaneni2024blindspots,gu2025surveyjudge}. Evaluator replacement has likewise been framed as a measurement-validity problem requiring workload-specific evaluation, bias probes, dependence estimates, and protocol audit trails \cite{yang2026judgechangesdoesmeasurement}. Selective verification adds another complication: when only suspected errors are adjudicated, the selection rule is part of the measurement instrument.

We study both sources of variation. The evidence comprises 2,550 formal outputs from three fixed Qwen snapshots and two fixed DeepSeek endpoints on 150 MATH and 150 ARC-Challenge questions. Qwen contributes the original 900-output measurement and a 450-output ARC cap control. DeepSeek contributes a balanced $2$ task $\times$ $2$ cap $\times$ $2$ model design with 1,200 outputs. Together these runs provide 1,050 matched question--model cap pairs: 450 Qwen ARC pairs and 600 DeepSeek pairs.

Three findings organize the paper. First, matched caps do not imply matched execution mixtures: at 2,048 tokens, the studied Qwen MATH configurations have 49/450 length-missing-final (LMF) outputs, whereas Qwen ARC has none. Second, the magnitude is configuration-specific: the studied DeepSeek MATH endpoints have 5/300 LMF at the same cap, 9.22 percentage points fewer than Qwen (95\% CI 5.22--13.67). At 8,192, no LMF is observed among the same DeepSeek pairs. Third, a targeted false-negative audit changes the DeepSeek-minus-Qwen MATH comparison from $-7.89$ points mechanically to $+6.89$ after unanimous panel adjudication. This reversal is descriptive of the audit pipeline, not a pure scorer effect, because verification coverage differs by configuration.

We make three contributions. First, we introduce a two-layer evaluation framework that separates pre-intervention execution evidence from scorer-dependent correctness. Second, we provide matched-budget measurements over 2,550 outputs and 1,050 paired cap comparisons, showing that execution case mix varies substantially across the studied task and model configurations. Third, we demonstrate through an auditable targeted-verification study that candidate-selection and aggregation policies can materially alter comparative accuracy conclusions. Together, these results establish execution state and verification policy as first-class components of language-model evaluation.

\begin{figure*}[t]
\centering
\includegraphics[width=\textwidth]{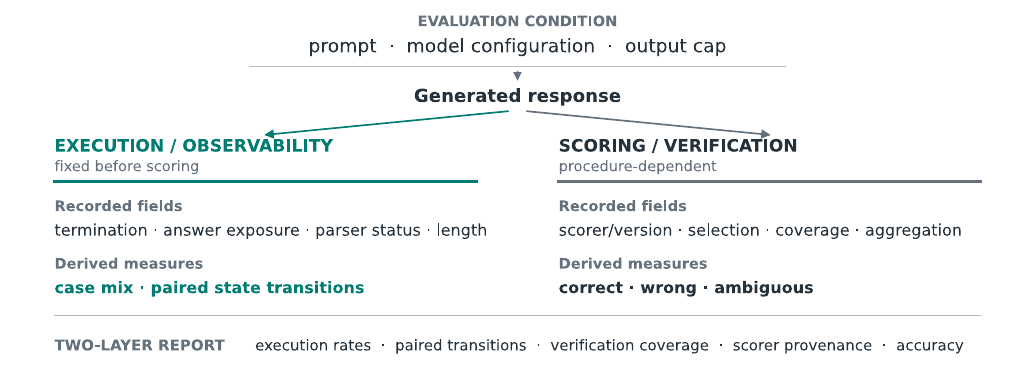}
\caption{Two-layer evaluation architecture. Execution evidence is frozen before correctness adjudication, while the scoring layer makes candidate selection, verification coverage, and aggregation policy explicit. Reporting both layers prevents accuracy from conflating response availability with grading decisions.}
\label{fig:architecture}
\end{figure*}

\section{Related Work}

\paragraph{Self-correction and test-time compute.}
Self-Refine iteratively revises outputs using self-generated feedback, while Reflexion uses verbal feedback and memory to improve subsequent attempts \cite{madaan2023selfrefine,shinn2023reflexion}; later analyses identify conditions under which intrinsic or externally supported correction helps or harms \cite{huang2024cannot,wu2024keycondition,kamoi2024when,pan2024automatically}. Chain-of-thought and self-consistency alter inference procedures \cite{wei2022chain,wang2023selfconsistency}, while adaptive-compute methods allocate additional inference based on difficulty or estimated benefit \cite{snell2025scaling,manvi2024adaptive}. Budget forcing explicitly shortens or extends reasoning traces \cite{muennighoff2025s1}. We measure the initial response states to which such interventions would be applied; we do not evaluate a recovery method.

\paragraph{Evaluation and failure analysis.}
Behavioral testing, measurement theory, and hypothesis-driven error analysis all caution against reducing evaluation to held-out accuracy \cite{ribeiro2020checklist,xiao2023evaluating,jones2022biases}. Benchmark validity and annotation reliability further constrain score interpretation \cite{bowman2021benchmarking}. ARC analyses characterize knowledge and reasoning demands \cite{boratko2018arc}, and semantic taxonomies organize generation defects \cite{huidrom2023consensustaxonomy}. Unlike semantic taxonomies, benchmark harnesses also expose execution evidence such as termination and parse success. We join that pre-intervention case mix with paired budget transitions and explicit scorer provenance.

\section{Measurement Design}

\subsection{Two Measurement Layers}

The \emph{execution/observability layer} records termination reason, whether an answer was exposed, parser status, and completion length. \textbf{LMF} denotes length termination without a parsed final answer; \textbf{LFP}, length termination with one; and \textbf{NLU}, non-length termination without a parseable answer. Answer availability is the presence of a parsed final answer regardless of its correctness. These endpoints are frozen before any LLM adjudication.

The \emph{scoring layer} assigns correct, wrong, or ambiguous under a stated scorer. For the mechanical scorer only, execution and correctness induce the familiar five-state projection: LMF, LFP, NLU, non-length wrong (NW), and non-length correct (NC). Targeted LLM labels are a correctness overlay on selected MATH candidates; they do not retroactively redefine parser status or create a second five-state partition. We therefore report termination, answer availability, and parsing separately from panel-accepted correctness.
Figure~\ref{fig:architecture} turns this separation into a minimum reporting protocol: execution, scoring, comparison, and provenance fields remain jointly visible without collapsing into one accuracy number.

\subsection{Questions, Models, and Prompts}

The fixed question sets contain 150 MATH training items, stratified across five difficulty levels \cite{hendrycks2021math}, and 150 ARC-Challenge validation items \cite{clark2018arc}. The Qwen variants are Qwen3.6-Flash, Qwen3.6-Plus, and Qwen3.7-Plus; the Qwen3 report provides family-level context rather than exact hosted-weight documentation \cite{yang2025qwen3}. The DeepSeek variants are the fixed hosted endpoints DeepSeek-V4-Flash and DeepSeek-V4-Pro.

The task prompts are held fixed across a task's caps and model families. Both place the required \texttt{FINAL:} line after reasoning or justification. MATH requests a derivation and final answer; ARC requests a concise justification followed by an option letter. A shared extractor takes the final \texttt{FINAL:} line or last balanced boxed expression. ARC additionally validates the extracted label against the question's option count. Thus answer placement is shared, but prompt wording, answer format, parser, content, and difficulty remain coupled to benchmark identity.

The Qwen Natural run contains MATH@2048 and ARC@8192 (900 outputs); a frozen sensitivity study reran the same 450 ARC question--model pairs at 2,048. DeepSeek crosses both tasks with caps 2,048 and 8,192 for each model, yielding 1,200 outputs and 600 pairs. Each target state was called once without result selection. Qwen used temperature zero, top-$p=1$, and disabled thinking, search, and tools. DeepSeek used the same task prompts, temperature zero, top-$p=1$, disabled thinking, and no tools or search. Hosted temperature-zero calls need not be bitwise deterministic, so small off-diagonal transitions are interpreted cautiously.

\begin{table*}[t]
\centering
\setlength{\tabcolsep}{4.0pt}
\begin{tabular}{lllrrrrr}
\toprule
Configuration & Task & Cap & $n$ & LMF & LFP & NLU & Answer available\\
\midrule
Qwen & MATH & 2048 & 450 & 49 & 1 & 3 & 398\\
Qwen & ARC & 2048 & 450 & 0 & 0 & 0 & 450\\
Qwen & ARC & 8192 & 450 & 0 & 0 & 0 & 450\\
\midrule
DeepSeek & MATH & 2048 & 300 & 5 & 0 & 2 & 293\\
DeepSeek & MATH & 8192 & 300 & 0 & 0 & 1 & 299\\
DeepSeek & ARC & 2048 & 300 & 0 & 0 & 0 & 300\\
DeepSeek & ARC & 8192 & 300 & 0 & 0 & 0 & 300\\
\bottomrule
\end{tabular}
\caption{Configuration--task--cap execution summaries. LMF, LFP, NLU, and answer availability are frozen before correctness adjudication.}
\label{tab:cells}
\end{table*}

\subsection{Scoring Provenance}

The mechanical scorer normalizes ARC labels by option count: numeric $1,\ldots,N$ maps to A through the $N$th option, while letters are uppercased and range-checked. MATH uses deterministic equivalence rules. A later audit locked 164 suspected MATH false negatives: 24 Qwen and 140 DeepSeek outputs. It did not re-score the complete sample.

Candidate selection was deterministic within each historical evidence boundary but not symmetric across configurations (Figure~\ref{fig:verification}a). The DeepSeek rule selected every successful MATH output that was mechanically non-correct and not LMF: 70/70 at 2,048 and 70/70 at 8,192. The earlier Qwen packet contained all 24 NW states, but not one LFP or three NLU states; complete Qwen response bodies were unavailable. Thus Qwen panel coverage is 24/77 of all mechanical non-correct outputs (31.2\%) and 24/28 of non-LMF mechanical non-correct outputs (85.7\%), versus 93.3\% and 100\% for DeepSeek@2048. Neither ARC sample was adjudicated.

Because protocol-compliant written human adjudication was unavailable, scorer sensitivity uses an availability-defined panel of the three judge lanes that completed all 264 blinded items: Codex GPT-5.6-sol, GLM-5.2, and LongCat-2.0. The originally planned Kimi lane remained incomplete and is excluded in full. The three-lane amendment was frozen after provider availability was known but before candidate-label aggregates were inspected. Judges saw the problem, reference answer, and candidate answer without model, family, cap, mechanical label, or peer judgments.

The packet mixes 100 mechanically certified controls (50 correct, 50 wrong) with the 164 formal candidates. All three judges scored 100/100 controls. Across all 264 items, 239 (90.53\%) received identical labels and Fleiss' $\kappa=0.851$. This is descriptive agreement among correlated LLM instruments, not a human gold standard.

We report four bases: mechanical; mechanical plus targeted Codex adjudication; mechanical plus targeted 3/3 unanimity; and mechanical plus targeted 2/3 majority. Codex is nested within the panel. Under unanimity, only three correct votes yield panel-accepted correct and only three wrong votes yield wrong; all other cases are ambiguous and count as wrong for accuracy. Under majority, two matching votes determine the label, with unresolved ambiguity again counted as wrong. Unanimity leaves 25/164 candidates ambiguous; majority leaves one. These are plausible audit policies, not four complete-sample gold scorers.

\subsection{Estimands and Uncertainty}

Within-family cap effects compare the same question and model at two caps. Cross-family comparisons at 2,048 use the same questions and task prompts, average equally across fixed models within each family, and remain associational. The DeepSeek task-by-cap interaction is
\[
(\mathrm{MATH}_{8192}-\mathrm{MATH}_{2048})
-
(\mathrm{ARC}_{8192}-\mathrm{ARC}_{2048}).
\]
Primary endpoints are LMF and answer availability for (i) Qwen MATH versus ARC at 2,048, (ii) the matched Qwen--DeepSeek MATH contrast, and (iii) the DeepSeek MATH cap contrast. Parser status, completion length, task-by-cap interaction, and all targeted correctness comparisons are secondary or exploratory. Confidence intervals are not adjusted for multiplicity.

All intervals use 10,000 deterministic PCG64 bootstrap replicates. Questions are the resampling units; all fixed-model observations and, for cap analyses, both caps travel with a sampled question. MATH and ARC are sampled independently for task contrasts. The seeds are 20260720 for the Qwen study and 20260722 for DeepSeek and cross-configuration analyses. Models are fixed cases, not independent population draws.

\begin{figure*}[t]
\centering
\includegraphics[width=\textwidth]{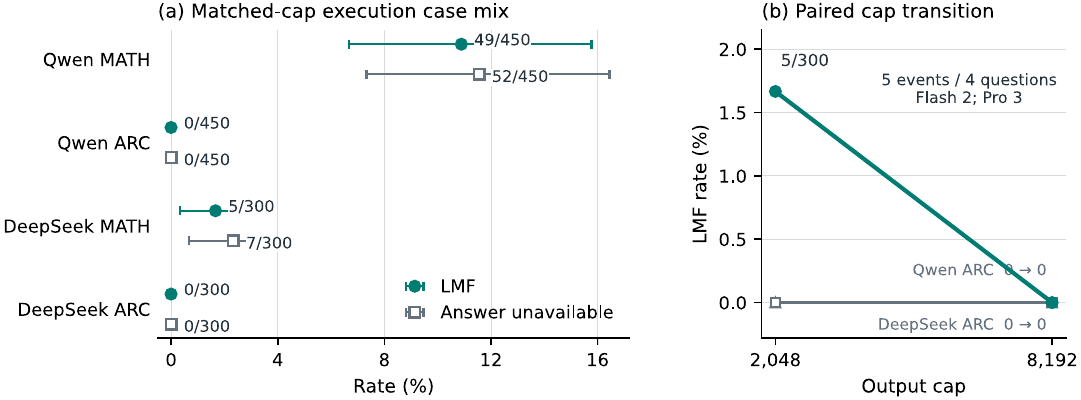}
\caption{Matched-cap execution mixtures and paired budget transitions. Panel (a) reports missing-final termination and answer unavailability before correctness adjudication; whiskers are question-clustered 95\% bootstrap intervals. Panel (b) shows within-question--model LMF changes across output caps. The studied Qwen MATH configurations present a substantially larger execution-failure case mix, while additional budget removes the five observed DeepSeek MATH LMF events. Degenerate zero intervals describe the fixed sample, not population zero-risk bounds.}
\label{fig:mixture}
\end{figure*}

\section{Results}

\subsection{Failure Mixtures Vary across Studied Configurations}

Figure~\ref{fig:mixture}a and Table~\ref{tab:cells} show the matched-cap result. Qwen MATH has 49/450 LMF (10.89\%), one LFP, and three NLU outputs, whereas Qwen ARC has none of these execution states. The contrast survives matching the configured cap, but characterizes the studied task--prompt setups rather than a causal task effect.

Qwen MATH answer availability is 398/450 (88.44\%), versus 450/450 for Qwen ARC. DeepSeek MATH has 5/300 LMF (1.67\%) and 293/300 answer-available outputs (97.67\%); DeepSeek ARC has zero LMF and 300/300 availability. The DeepSeek-minus-Qwen MATH LMF difference is $-9.22$ percentage points (95\% CI $[-13.67,-5.22]$), and answer availability is $+9.22$ points ($[5.22,13.67]$). The model-level LMF counts are 16, 16, and 17 for the three Qwen variants, and 2 and 3 for DeepSeek Flash and Pro; no single endpoint drives the direction. These are fixed-configuration associations, not family-level causal effects.

\subsection{Paired Cap Changes}

The Qwen ARC control contains 450 pairs. Lowering the cap from 8,192 to 2,048 produces no LMF or answer-availability change: 441 pairs remain NC, seven remain NW, and two move from NW to NC under normalized mechanical scoring. This small favorable accuracy difference ($+0.44$ points, 95\% CI $[0,1.11]$) is not evidence that a smaller cap improves accuracy.

For DeepSeek MATH, Figure~\ref{fig:mixture}b shows LMF changing from 5/300 at 2,048 to 0/300 at 8,192, a paired difference of $-1.67$ points (95\% bootstrap CI $[-3.67,-0.33]$). The five events occur across four questions (two in Flash and three in Pro); under mechanical scoring, three transition to NC and two to NW. With only five one-way discordances, the exact two-sided sign/McNemar probability is $p=0.0625$; the fixed-question bootstrap interval is not an equivalent population-null test. Answer availability increases by $+2.00$ points ($[0.33,4.00]$). DeepSeek ARC remains at zero LMF and full availability at both caps. The MATH-minus-ARC task-by-cap interaction is $-1.67$ points for LMF ($[-3.67,-0.33]$) and $+2.00$ for availability ($[0.33,4.00]$).

\begin{figure*}[t]
\centering
\includegraphics[width=\textwidth]{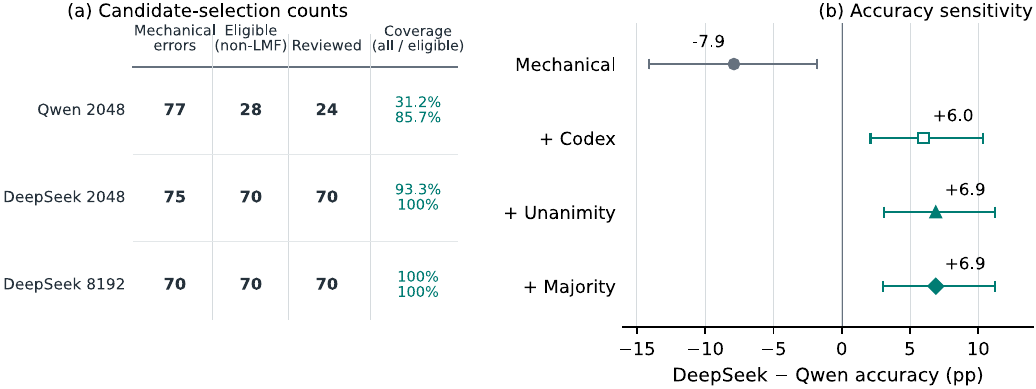}
\caption{Targeted-verification pipeline and exploratory sensitivity. Panel (a) shows configuration-specific candidate coverage. Panel (b) reports the matched MATH@2048 comparison under the mechanical baseline and three targeted verification policies; whiskers are question-clustered 95\% bootstrap intervals. Because coverage differs across configurations, the shift reflects candidate selection and scoring jointly.}
\label{fig:verification}
\end{figure*}

\subsection{Targeted Verification Alters the Exploratory Comparison}

Figure~\ref{fig:verification}b gives the matched MATH@2048 contrast. Mechanical scoring yields a DeepSeek-minus-Qwen difference of $-7.89$ points (95\% CI $[-14.11,-1.78]$). Mechanical plus targeted Codex adjudication yields $+6.00$ ($[2.11,10.33]$); targeted 3/3 yields $+6.89$ ($[3.11,11.22]$); and targeted 2/3 yields $+6.89$ ($[3.00,11.22]$). Every estimate uses the same questions, cap, and within-configuration equal-model average, but verification opportunity is not matched: 24 Qwen and 70 DeepSeek@2048 outputs were panel candidates. The reversal therefore demonstrates sensitivity to this \emph{targeted audit pipeline}, jointly reflecting scorer and selection policy; it does not identify a general grader effect.

ARC@2048 does not reverse: every basis gives DeepSeek minus Qwen $-1.78$ points (95\% CI $[-3.33,-0.44]$), because the adjudication candidates are MATH responses. Within DeepSeek, strict correctness changes by $+1.33$ points for MATH ($[-1.67,4.33]$) and $+0.33$ for ARC ($[0,1.00]$) when raising the cap. These small effects do not alter the termination result.

The panel is highly consistent but aggregation still matters. Among 164 candidates, unanimity assigns 118 panel-accepted correct, 21 wrong, and 25 ambiguous; majority assigns 134, 29, and one. Sixteen candidates change accuracy status solely through aggregation. Agreement supports stability of the instrument, not a unique truth.

\section{Implications and Limitations}

Pre-intervention case mix is an evaluation variable. A recovery method evaluated on Qwen MATH would encounter many missing answers at a token boundary; the same protocol on DeepSeek MATH or ARC would encounter a different mixture. Reporting only post-intervention accuracy makes method performance inseparable from this starting distribution. A minimal evaluation should report unconditional response-state rates before intervention and state transitions afterward.

Verification policy is a second evaluation variable. The termination results do not consult a correctness judge, whereas the matched MATH accuracy comparison changes after targeted adjudication. Because candidate opportunity differs across configurations, the result implicates the combined selection-and-scoring pipeline rather than grader choice alone. Benchmark reports should distinguish execution evidence from correctness, disclose what was sent for verification, and retain mechanical baselines.

The evidence has several limits. It covers two tasks, three Qwen snapshots, and two DeepSeek endpoints; it does not support population claims about model families, code, open-ended QA, or dialogue. Cross-configuration associations bundle architecture, scale, serving stack, and generation behavior. Cross-task associations bundle content, difficulty, prompt wording, answer format, and parser, although both prompts place the answer last and the cap is matched. MATH uses training-split questions that may have appeared in model pretraining; the study measures response states rather than uncontaminated capability.

The three-judge panel is an LLM-based instrument, not a human gold standard. Shared training data and correlated reasoning errors can produce agreement without correctness. Its membership was determined after Kimi provider availability was observed, although before candidate aggregates were inspected. The controls establish performance on mechanically certified equivalences, not all contested mathematics. Codex is nested within the panel. Most importantly, selective, differential verification prevents interpreting panel-accepted accuracy as complete-sample truth.

The Qwen Natural ledger did not retain complete response bodies; its later scoring analysis relies on parsed answers and response hashes rather than independent raw-text reparse. DeepSeek and ARC-sensitivity evidence retain complete bodies. Scoring audits were initiated after observing the initial results. We therefore make termination and answer availability primary, present scorer-sensitive correctness as sensitivity analysis, and avoid claims of a validated recovery mechanism.

\section{Reproducibility and Responsible Reporting}

The anonymous artifact contains frozen sample identifiers, prompt and model bindings, aggregate evidence, scorer definitions, and offline analysis code. It now also releases 164 anonymized item-level mechanical and three-judge labels, strict and majority aggregates, candidate membership, model/task/cap binding, and a candidate-coverage audit. Questions, answers, reasons, provider identifiers, credentials, account data, and local paths remain excluded. One offline builder reconstructs the matched-slice statistics and figures and records source hashes. Historical ledgers and amendments remain unchanged.

The datasets contain benchmark questions rather than personal data. Hosted outputs are analyzed as system behavior, not attributed to people. No Stage B recovery experiment is analyzed.

\section{Conclusion}

Across 2,550 outputs, accuracy hides two forms of measurement variation. Execution mixtures differ across the studied task, cap, and model configurations: Qwen MATH exhibits substantially more missing-final length termination than DeepSeek MATH, while ARC exhibits none. Among 300 paired DeepSeek MATH observations, five LMF events occur at 2,048 and none at 8,192. Correctness estimates also change under coverage-audited targeted verification. Evaluations of test-time methods should report execution states, verification coverage, and scorer provenance rather than treating every zero score as equivalent.

\bibliographystyle{IEEEtran}
\bibliography{references}
\end{document}